\documentclass[sigconf]{acmart}
\setcopyright{none}
\renewcommand\footnotetextcopyrightpermission[1]{}
\acmYear{2027}
\copyrightyear{2027}
\acmConference[WWW '27]{The Web Conference 2027}{May 10--14, 2027}{Dublin, Ireland}
\acmISBN{}
\acmDOI{}

\usepackage{amsmath}
\IfFileExists{algorithmic.sty}{\usepackage{algorithmic}}{\usepackage{algorithmic-local}}
\IfFileExists{algorithm.sty}{\usepackage{algorithm}}{\usepackage{algorithm-local}}
\usepackage{graphicx}
\usepackage{xcolor}
\usepackage{multirow}
\usepackage{adjustbox}
\usepackage{enumitem}
\renewcommand{\arraystretch}{1.05}

\newcommand{\R}{\mathbb{R}}

\title[CoGSeq-LTV]{CoGSeq-LTV: Scalable Missing-Aware Graph-Sequence Modeling for Multi-Business Transaction Analytics}
\author{Ziyang Su}
\affiliation{%
  \department{School of Computer Science}
  \institution{Shanghai Jiao Tong University}
  \city{Shanghai}
  \country{China}}
\email{suziyang@sjtu.edu.cn}

\author{Yao Zhou}
\affiliation{%
  \institution{Tencent WeChat Pay Research and Development Department}
  \city{Shenzhen}
  \country{China}}
\email{yaozzhou@tencent.com}

\author{Kelin Liu}
\affiliation{%
  \institution{Tencent WeChat Pay Research and Development Department}
  \city{Shenzhen}
  \country{China}}
\email{clingliu@tencent.com}

\author{Guanjie Zheng}
\affiliation{%
  \department{School of Computer Science}
  \institution{Shanghai Jiao Tong University}
  \city{Shanghai}
  \country{China}}
\email{gjzheng@sjtu.edu.cn}

\author{Linghe Kong}
\affiliation{%
  \department{School of Computer Science}
  \institution{Shanghai Jiao Tong University}
  \city{Shanghai}
  \country{China}}
\email{linghe.kong@sjtu.edu.cn}

\begin{document}
\begin{abstract}
Large-scale super-apps generate heterogeneous transaction logs in which users interact with multiple businesses, item identifiers are only partially available, and future value targets are sparse, heavy-tailed, and cumulative over time. We study multi-business, multi-horizon value prediction as a data-intensive analytics problem over such logs, rather than as a single-table task. We propose \textbf{CoGSeq-LTV}, a contrastive graph–sequence model that integrates three complementary modeling views: (i) a batch-induced user-item graph that provides collaborative signals without materializing a global graph and remains applicable when item identifiers are missing through reserved unknown-item handling, (ii) business-aware sequence encoders that preserve recency, inter-event gaps, payment context, and goods-table availability, and (iii) cross-view contrastive alignment that transfers user-level intent across businesses while retaining business-specific temporal patterns. To handle zero-inflated and long-tailed cumulative targets, CoGSeq-LTV adopts a zero-inflated mixture log-normal head with MAE-aligned point supervision and validation-calibrated zero-aware inference. Experiments on the BiBiSheng transaction-log dataset show that CoGSeq-LTV improves over representative LTV baselines across businesses and horizons, including under a later out-of-time evaluation window.
\end{abstract}

\ccsdesc[500]{Information systems~Data mining}
\ccsdesc[300]{Information systems~Data analytics}
\ccsdesc[300]{Computing methodologies~Machine learning}
\keywords{Graph-sequence learning, Lifetime value, Missing-aware modeling}
\maketitle

\section{Introduction}

Large consumer platforms are increasingly organized as super-apps that host multiple businesses in a shared user ecosystem. Their analytical pipelines must reason over heterogeneous transaction logs: the same user may purchase across businesses with different rhythms, monetary scales, item namespaces, and metadata coverage. In this setting, customer lifetime value (LTV) is not merely a marketing score, but a downstream analytical target derived from large-scale temporal logs. Reliable value prediction requires handling partially missing item identifiers, cross-business behavioral coupling, zero-inflated and heavy-tailed outcomes, and strict temporal separation between observed features and future labels \cite{berger1998customer,schmittlein1987counting,vanderveld2016engagement,wang2019deep,li2022billion}.

We study \emph{multi-business, multi-horizon} LTV prediction in such a data-intensive setting. Each user may transact in multiple businesses---e-commerce and food delivery in our dataset---producing parallel event streams with different spending scales, interaction frequencies, and item-metadata quality. For each business, the platform requires a vector of cumulative future values over multiple horizons (e.g., 7/14/30/45 days), so that short-term intervention and longer-term resource planning can be coordinated. This makes the task substantially harder than standard single-business CLV prediction because the model must jointly preserve temporal order, exploit cross-business signals, and respect the cumulative nature of the targets.

Several data characteristics make the problem challenging. First, LTV labels are typically \emph{zero-inflated} and \emph{heavy-tailed}: many users spend nothing within a given horizon, whereas a small set of ``whale'' users contributes a disproportionate amount of revenue \cite{wang2019deep,li2022billion,zhang2023out}. Second, user value is inherently temporal. Transaction order, recency, and irregular inter-event gaps carry predictive signal that can be lost in coarse aggregate tables \cite{vaswani2017attention,kang2018sasrec}. Third, the two businesses are correlated but not identical: they reflect partially shared user intent and activity level, yet their spending patterns and temporal dynamics differ. Fourth, item information is heterogeneously available across channels. Some transactions contain valid \texttt{goods\_id} and matched goods-side attributes, while others have missing identifiers or no goods-table match due to cross-platform navigation. This systematic missingness creates a practical data-quality challenge for methods that assume uniformly available item metadata.

Existing LTV methods have made important progress on distributional modeling. ZILN-style objectives separate zero-purchase probability from positive-value magnitude under a log-normal assumption \cite{wang2019deep}, and recent systems introduce expert routing, adaptive distributions, or multi-distribution decomposition to better fit long-tailed user-value regimes \cite{li2022billion,weng2024optdist,liu2024mdan,zhang2023out}. In parallel, mixture density modeling provides a flexible mechanism for representing multi-regime outcomes \cite{bishop1994mdn}, and multi-task backbones such as MMoE and PLE provide useful inductive biases for related prediction targets \cite{ma2018mmoe,tang2020ple}. However, most existing LTV solutions are still centered on single-business feature tables or coarse aggregates. They do not explicitly combine transaction-level temporal modeling, cross-business coupling, and user--item relational structure under partially missing item information. Sequence models can capture fine-grained temporal dependencies \cite{vaswani2017attention,kang2018sasrec}, graph neural networks can encode interaction structure \cite{kipf2017semi,hu2020heterogeneous}, and contrastive learning is effective for exploiting correlated views under sparse supervision \cite{oord2018representation}; the open question is how to integrate these components into a scalable and missing-aware analytics pipeline for heterogeneous transaction logs.

To address this question, we propose \textbf{CoGSeq-LTV}, an end-to-end graph-sequence framework for multi-business transaction-log analytics. The design focuses on two principles. First, the relational component should be lightweight enough for mini-batch training and should not require global graph materialization. We therefore construct a \emph{batch-induced user--item bipartite graph} that enriches user and item representations locally while sharing item embeddings globally across batches. Missing item identifiers are handled explicitly through reserved unknown-item representations and goods-table match indicators, rather than by assuming complete item attributes. Second, the temporal component should keep business-specific dynamics while enabling cross-business transfer. CoGSeq-LTV builds business-aware event tokens from graph-enhanced item embeddings, goods numerical features, amount, recency, inter-event gaps, and match indicators, then feeds the two business lanes into separate Transformer encoders. A contrastive objective aligns the e-commerce, food-delivery, and graph views of the same user to exploit shared intent without collapsing business-specific patterns.

For the prediction layer, CoGSeq-LTV adopts a \emph{zero-inflated mixture log-normal} (ZIMLN) head. The head separates zero outcomes from a flexible mixture model for positive values, and an MAE-aligned auxiliary term connects the likelihood objective to the evaluation metric. Because the targets are cumulative across horizons, the problem definition explicitly records the nondecreasing label structure; in this version, we use validation-calibrated zero-aware inference for point prediction while leaving hard monotonic output constraints as a natural extension. We also specify the data protocol and temporal cutoffs used to prevent feature-label leakage.

We evaluate CoGSeq-LTV on the BiBiSheng dataset, a large-scale real-world transaction-log dataset with two businesses and heterogeneous item availability. Experiments show that CoGSeq-LTV consistently outperforms representative CLTV/LTV baselines across both businesses and all prediction horizons under MAE and WAPE. Its advantage also persists in a later out-of-time evaluation window, suggesting stronger robustness under temporal distribution shift rather than gains tied to a single evaluation period.

Our main contributions are summarized as follows:
\begin{itemize}
    \item We formulate multi-business, multi-horizon LTV prediction as a heterogeneous transaction-log analytics problem with realistic data-engineering constraints, including systematic item-metadata missingness, temporal cutoffs, zero-inflated heavy-tailed labels, and cumulative targets.
    \item We propose \textbf{CoGSeq-LTV}, a missing-aware graph-sequence architecture that combines batch-induced user--item propagation, business-specific temporal encoders, and cross-view contrastive alignment without requiring global graph materialization.
    \item We introduce a distribution-aware prediction design based on ZIMLN, MAE-aligned auxiliary supervision, and zero-aware inference to better handle sparse and long-tailed multi-horizon value targets.
    \item We provide empirical evidence on the BiBiSheng transaction-log dataset, including main accuracy, ablation, user-frequency analysis, and out-of-time robustness evaluation, showing consistent improvements over representative baselines.
\end{itemize}

\section{Related Work}
\label{sec:related}

\subsection{LTV/CLV Prediction}
Classical CLV models use probabilistic repeat-purchase assumptions, such as BTYD formulations~\cite{schmittlein1987counting}, or interpretable RFM segmentation~\cite{fader2005rfm}; these approaches do not directly exploit the high-dimensional signals available in transaction logs~\cite{berger1998customer}. Industrial systems therefore increasingly learn propensity and value jointly or in two stages~\cite{vanderveld2016engagement}. Recent methods model the target distribution explicitly, including zero-inflated log-normal objectives~\cite{wang2019deep} and large-scale multi-horizon architectures~\cite{li2022billion}. CoGSeq-LTV extends this line to a setting with two correlated businesses, event-level histories, and systematic item-metadata gaps.

\subsection{Distribution-aware Multi-horizon Learning}
LTV labels are zero-inflated and heavy-tailed, so a single point-regression objective can be dominated by a small number of whale users. Mixture density models represent multiple positive-value regimes~\cite{bishop1994mdn}, while OptDist and MDAN adapt distribution selection or multi-distribution learning to improve robustness~\cite{weng2024optdist,liu2024mdan}. The four horizons also form a related multi-task problem; MMoE and PLE provide shared--specific parameterizations for such settings~\cite{ma2018mmoe,tang2020ple}. Our model combines a zero-inflated mixture head with relational and temporal representations instead of treating horizons as independent regressions.

\subsection{Graph and Temporal Analytics}
GCN, relational GCN, and heterogeneous graph transformers provide general message-passing mechanisms for interaction data~\cite{kipf2017semi,schlichtkrull2018modeling,hu2020heterogeneous}, but global graph construction and neighbor sampling can be costly at platform scale. CoGSeq-LTV builds a batch-induced user--item graph, retaining collaborative propagation while keeping each training graph small. Transformer encoders and sequential recommenders such as SASRec capture long-range event dependencies~\cite{vaswani2017attention,kang2018sasrec}; here, recency, inter-event gaps, amount, business identity, and metadata availability are encoded with each event because order alone does not resolve irregular and partially observed transaction histories.

\subsection{Missingness and Self-supervision}
Cross-channel integration often makes item identifiers systematically missing rather than randomly absent. We therefore use reserved unknown-item states, zero-filled goods features, and an explicit match indicator so the encoder can distinguish matched items, unmatched-but-hashable items, and completely unknown events. Contrastive learning is a natural complement when labels are sparse~\cite{oord2018representation}: the same user supplies e-commerce, food-delivery, and graph views. Our cross-view objective aligns shared intent across these views while leaving the supervised head free to retain business- and horizon-specific information.

\section{Preliminaries and Data Protocol}
\label{sec:prelim}

\subsection{Problem Definition}
\label{subsec:problem}

We study multi-business, multi-horizon lifetime value (LTV) prediction from transaction-level logs.
Let $\mathcal{U}$ denote the user set, $\mathcal{B}=\{E,F\}$ denote two businesses, and $\mathcal{H}=\{7,14,30,45\}$ denote the prediction horizons (in days).
We fix a prediction start time $t_0$ (Oct.~1st in our main experiments).
All observable features are restricted to information available before $t_0$, and all labels are computed from transactions after $t_0$.

For each user $u\in\mathcal{U}$ and business $b\in\mathcal{B}$, the historical transactions before $t_0$ form a variable-length sequence
\begin{equation}
\mathcal{S}_u^b=\Bigl\{\bigl(t_{u,i}^b,\,a_{u,i}^b,\,g_{u,i}^b\bigr)\Bigr\}_{i=1}^{n_u^b},
\qquad t_{u,i}^b<t_0,
\label{eq:seq_def}
\end{equation}
where $t_{u,i}^b$ is the transaction time, $a_{u,i}^b\in\R_{\ge 0}$ is the payment amount, and $g_{u,i}^b$ is the item identifier.
Some transactions may have missing item identifiers, denoted by $g_{u,i}^b=\varnothing$.
We use
\begin{equation}
m_{u,i}^b=\mathbf{1}\!\left(g_{u,i}^b\neq \varnothing\right)
\label{eq:item_missing_indicator}
\end{equation}
to indicate whether item information is available.

Each user is also associated with a static feature vector $\mathbf{s}_u\in\R^{d_s}$ whose values are generated from pre-$t_0$ snapshots.
The full input for user $u$ is
\begin{equation}
\mathbf{x}_u=\Bigl(\mathbf{s}_u,\,\mathcal{S}_u^E,\,\mathcal{S}_u^F\Bigr).
\label{eq:input_def}
\end{equation}

For each business--horizon pair $(b,h)\in\mathcal{B}\times\mathcal{H}$, the target is the cumulative spending in the future window $[t_0,t_0+h)$:
\begin{equation}
y_u^{b,h}=\sum_{i:\, t_0 \le t_{u,i}^b < t_0+h} a_{u,i}^b.
\label{eq:ltv_def}
\end{equation}
We stack the resulting eight targets into $\mathbf{y}_u\in\R_{\ge 0}^{8}$.
Given a dataset $\mathcal{D}=\{(\mathbf{x}_u,\mathbf{y}_u)\}_{u=1}^{N}$, we learn a predictor $f_\theta$ such that $\widehat{\mathbf{y}}_u=f_\theta(\mathbf{x}_u)$.
We evaluate prediction quality by mean absolute error (MAE), averaged over users, businesses, and horizons:
\begin{equation}
\mathrm{MAE}
=
\frac{1}{N|\mathcal{B}||\mathcal{H}|}
\sum_{u=1}^{N}\sum_{b\in\mathcal{B}}\sum_{h\in\mathcal{H}}
\left|\widehat{y}_u^{b,h}-y_u^{b,h}\right|.
\label{eq:mae}
\end{equation}

\subsection{Data Protocol and Leakage Control}
\label{subsec:data_protocol}

The prediction protocol follows an anchor-time design. For a given anchor $t_0$, model inputs include only static user features, item/goods features, and transaction events whose timestamps are strictly earlier than $t_0$. Future transactions in $[t_0,t_0+h)$ are used only to construct labels. Missing identifiers use the reserved unknown id; identifiable but unmatched items retain their hashed ids with zero-filled attributes; the match flag distinguishes both states from matched items. This protocol prevents the model from using future item popularity, future user activity, or post-anchor goods statistics.

Train/validation/test partitions are performed at the user level so that the same user does not appear in more than one split. Hyperparameters, zero-aware inference thresholds, target caps, and checkpoint selection are determined using only the training and validation splits. The held-out test split is used once for final reporting. The out-of-time robustness experiment keeps the trained model and preprocessing rules fixed and evaluates on a later anchor window, which tests temporal generalization without updating model parameters using future data.

\subsection{Target Characteristics}
\label{subsec:ziln_prelim}

The LTV targets in our setting have two properties that are important for model design.
First, they are \emph{zero-inflated} and \emph{heavy-tailed}: many users have zero future spending, while positive values vary substantially and a small fraction of \emph{whale} users contribute extremely large amounts.
Second, the targets are cumulative across horizons.
For the same business, if $h_1<h_2$, then $y_u^{b,h_1}\le y_u^{b,h_2}$.
Equivalently, multi-horizon LTV can be viewed as cumulative sums of nonnegative increments, e.g., $\Delta y_u^{b,h_j}=y_u^{b,h_j}-y_u^{b,h_{j-1}}\ge 0$.

These properties motivate the distributional prediction head introduced in Sec.~\ref{subsec:zimln}.

\section{Methodology}
\label{sec:method}

\subsection{Overview}
\begin{figure*}[t]
    \centering
    \includegraphics[width=\linewidth]{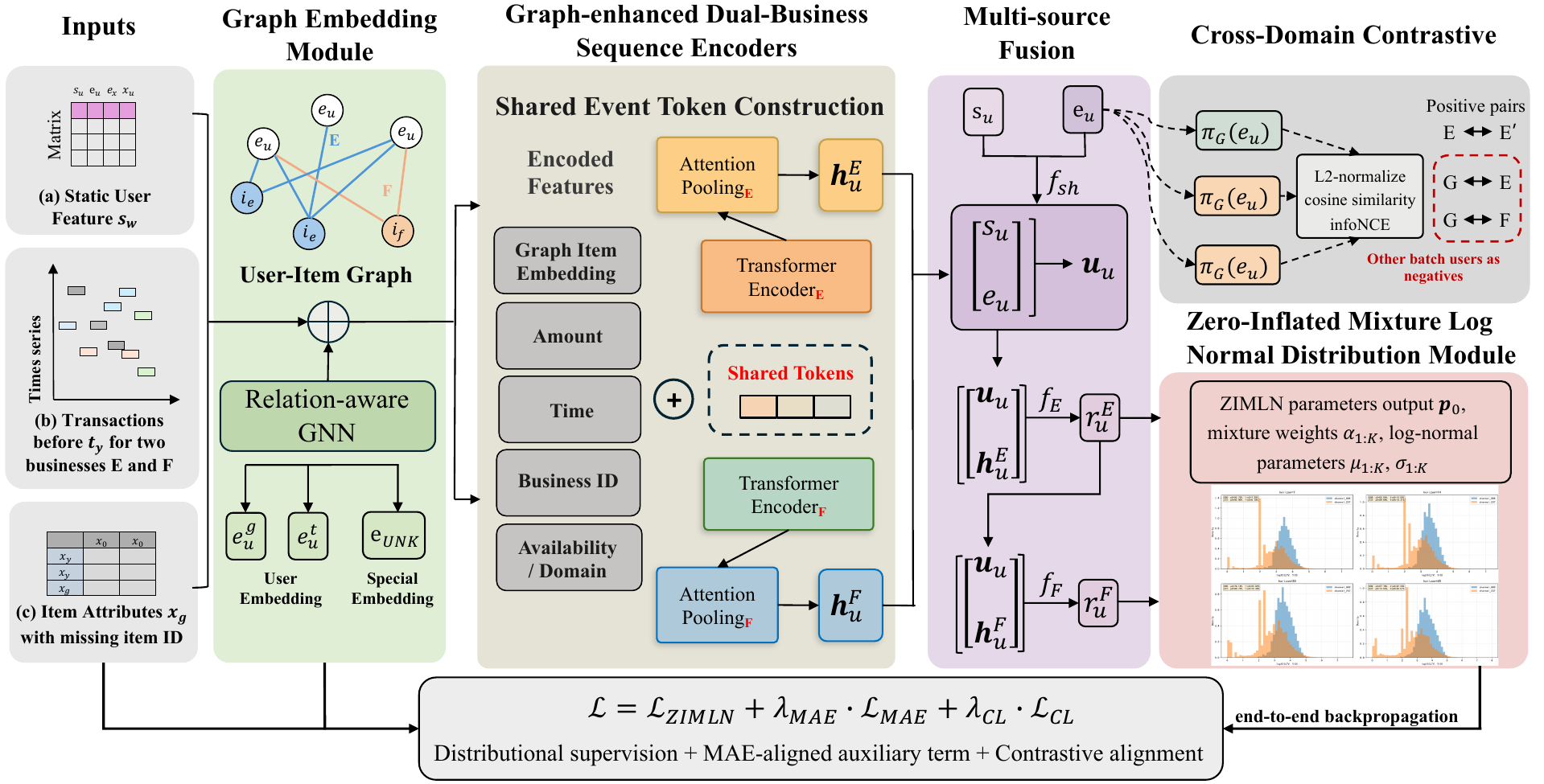}
    \caption{The overall architecture of CoGSeq-LTV.}
    \Description{A block diagram showing a shared user representation, a batch-induced user-item graph encoder, two business-specific transaction sequence encoders, cross-business contrastive alignment, and a zero-inflated mixture log-normal prediction head.}
    \label{fig:overall}
\end{figure*}

CoGSeq-LTV combines four views of a user's pre-anchor history: a mini-batch user--item graph, business-specific transaction sequences, cross-business contrastive alignment, and a zero-inflated mixture log-normal (ZIMLN) prediction head. The graph is built only from the current batch, which supplies collaborative context without materializing a platform-scale graph. Missing item identifiers use an explicit unknown-item state and a match flag, so unavailable metadata is not confused with a valid item whose numerical features happen to be small. Figure~\ref{fig:overall} summarizes the data flow.

Static user features produce a shared state, which is enhanced by graph propagation and fused with a sequence summary for each business:
\begin{equation}
\mathbf{u}_u^{\mathrm{base}}=\mathrm{MLP}_{\mathrm{user}}(\mathbf{s}_u),\qquad
\mathbf{u}_u=\mathbf{u}_u^{\mathrm{base}}+W_g\mathbf{g}_u,
\label{eq:user_fusion}
\end{equation}
\begin{equation}
\mathbf{r}_u^b=f_b([\mathbf{h}_u^b;\mathbf{u}_u]),\qquad b\in\{E,F\}.
\label{eq:biz_user_repr}
\end{equation}
The user MLP maps static features to the shared hidden space, while each $f_b$ is a business-specific feed-forward layer with ReLU and dropout. The same shared state is therefore available to both businesses, but the sequence summary and prediction head remain business-specific. Each head predicts all four horizons through a shared hidden layer and horizon-specific outputs, allowing related windows to exchange statistical strength without forcing their target distributions to be identical.

\subsection{Batch-Induced User--Item Graph}
For an event with normalized item key $\rho_{u,i}^b$, the shared item index is
\begin{equation}
\iota_{u,i}^b=\begin{cases}2,&\rho_{u,i}^b=\varnothing,\\
3+\operatorname{hash}(\rho_{u,i}^b)\bmod M,&\text{otherwise},
\end{cases}
\label{eq:item_hash}
\end{equation}
where $M=200{,}000$ and ids 0, 1, and 2 are reserved for padding, \texttt{NO\_EVENT}, and unknown items. For a mini-batch, we build a bipartite graph between users and the unique item ids in their two histories. Padding and \texttt{NO\_EVENT} are excluded; the unknown-item node remains available for missing identifiers.

For e-commerce, the normalized key is the channel--goods pair; food-delivery events use the shared food channel namespace. The deterministic hash gives the same item index whenever a key reappears, while the reserved unknown index is shared by events whose identifier is absent. The graph is intentionally local to the mini-batch: it approximates collaborative propagation without storing a platform-wide adjacency structure. Because the item embedding table is shared across batches, repeated updates still accumulate information about frequently observed items.

User and item nodes are initialized as $\mathbf{q}_u^{(0)}=W_U\mathbf{u}_u^{\mathrm{base}}$ and $\mathbf{q}_i^{(0)}=\mathrm{Emb}(i)$. One bidirectional mean-aggregation layer gives
\begin{equation}
\begin{aligned}
\mathbf{q}_u^{(1)}&=\operatorname{LN}\!\left(\mathbf{q}_u^{(0)}+\phi_U\!\left(\frac{1}{|\mathcal{N}(u)|}\sum_{i\in\mathcal{N}(u)}\mathbf{q}_i^{(0)}\right)\right),\\
\mathbf{q}_i^{(1)}&=\operatorname{LN}\!\left(\mathbf{q}_i^{(0)}+\phi_I\!\left(\frac{1}{|\mathcal{N}(i)|}\sum_{u\in\mathcal{N}(i)}\mathbf{q}_u^{(0)}\right)\right).
\end{aligned}
\label{eq:batch_graph_update}
\end{equation}
The resulting $\mathbf{g}_u=\mathbf{q}_u^{(1)}$ and $\mathbf{g}_{u,i}^b=\mathbf{q}_{\iota_{u,i}^b}^{(1)}$ enrich both the shared user stream and event tokens. Because the item table is shared across batches, local graph instances accumulate global item information during training. The graph is rebuilt for every mini-batch, so its working memory grows with the users and unique items in that batch rather than with the full platform graph. This makes the relational signal compatible with ordinary mini-batch optimization and avoids a separate graph-sampling pipeline.

\subsection{Dual-Business Sequence Encoder}
For each business $b$, the pre-anchor history is a variable-length sequence $\mathcal{S}_u^b=\{(t_{u,i}^b,a_{u,i}^b,g_{u,i}^b)\}_{i=1}^{n_u^b}$. Each event token combines the graph-enhanced item state, standardized goods features, log amount, recency, inter-event gap, and the goods-table match flag:
\begin{equation}
\mathbf{c}_{u,i}^b=\operatorname{MLP}_{\mathrm{ctx}}^b([\log(1+a_{u,i}^b);\Delta_{u,i}^{\tau};\Delta_{u,i}^{t};d_{u,i}^b]),
\end{equation}
\begin{equation}
\mathbf{z}_{u,i}^b=\operatorname{LN}\!\left(W_{\mathrm{out}}^b[\mathbf{g}_{u,i}^b;W_{\mathrm{num}}^b\mathbf{x}_{u,i}^{b,\mathrm{num}};\mathbf{c}_{u,i}^b]+\mathbf{b}_{\mathrm{biz}}^b\right).
\label{eq:event_embed}
\end{equation}
A missing history receives a reserved \texttt{NO\_EVENT} token, while unavailable goods metadata is zero-filled and marked by $d_{u,i}^b=0$. Thus the event representation distinguishes a matched goods record, an observed transaction without a goods-table join, and an event whose item identifier is missing entirely. The continuous context features are computed before padding and are standardized using training users only; padding positions are masked in both self-attention and pooling. The two business lanes use separate Transformer parameters:
\begin{equation}
\mathbf{H}_u^b=\operatorname{Transformer}_b([\mathbf{z}_{u,1}^b,\ldots,\mathbf{z}_{u,n_u^b}^b]^\top),\qquad
\mathbf{h}_u^b=\sum_{i:\mathrm{valid}}\alpha_{u,i}^b\mathbf{H}_{u,i}^b,
\label{eq:seq_encoder}
\end{equation}
where $\alpha_{u,i}^b$ is a masked softmax of a learned score. Continuous recency and inter-event gaps provide temporal position information without a discrete position embedding. The pooling weights are computed only on non-padding events,
\begin{equation}
 s_{u,i}^b=\mathbf{w}_b^\top\mathbf{H}_{u,i}^b,\qquad
 \alpha_{u,i}^b=\frac{\exp(s_{u,i}^b)}{\sum_{j:\mathrm{valid}}\exp(s_{u,j}^b)},
 \label{eq:attn_pool_short}
\end{equation}
so recent or unusually informative transactions can receive more weight than routine events. The explicit match flag is retained even when goods features are zero-filled; this separates a valid item with small numerical attributes from an unavailable goods-table join. A missing business history instead receives the reserved \texttt{NO\_EVENT} token, which keeps both business lanes well-defined for every user.

\subsection{Cross-View Contrastive Alignment}
The e-commerce, food-delivery, and graph views are projected into a shared contrastive space. Same-user pairs are positives and other users in the mini-batch are negatives. For an ordered view pair $A\rightarrow B$,
\begin{equation}
\mathcal{L}_{A\rightarrow B}=-\frac{1}{|\mathcal{U}_B|}\sum_{u\in\mathcal{U}_B}\log\frac{\exp(s_{uu}^{A,B})}{\sum_{v\in\mathcal{U}_B}\exp(s_{uv}^{A,B})},
\label{eq:infonce}
\end{equation}
where $s_{uv}^{A,B}=\bar{\mathbf{z}}_u^{A\top}\bar{\mathbf{z}}_v^B/\tau$. The symmetric objective is
\begin{equation}
\mathcal{L}_{\mathrm{CL}}=\mathcal{L}_{E\leftrightarrow F}+\lambda_G(\mathcal{L}_{G\leftrightarrow E}+\mathcal{L}_{G\leftrightarrow F}).
\label{eq:cl_total}
\end{equation}
This transfers shared user intent across businesses while retaining business-specific temporal patterns and avoiding a memory bank. In implementation, each view is first mapped by a two-layer projection head to a common contrastive dimension and then $\ell_2$-normalized. Same-user pairs are positives, whereas the other users in the current mini-batch are negatives. The cross-business term transfers propensity signals when one business history is sparse; the graph--business terms regularize the collaborative and sequential views without requiring a momentum encoder or an external memory bank.

\subsection{ZIMLN Prediction Head and Training}
\label{subsec:zimln}
LTV targets are nonnegative, zero-inflated, and heavy-tailed. For each business--horizon pair $(b,h)$, the positive part is a $K$-component mixture of LogNormals:
\begin{equation}
p_+^{b,h}(y)=\sum_{k=1}^{K}\alpha_k^{b,h}\operatorname{LogNormal}(y;\mu_k^{b,h},\sigma_k^{b,h}),
\label{eq:zimln_pos}
\end{equation}
with full conditional distribution
\begin{equation}
p(y\mid\mathbf{r}_u^b)=p_0^{b,h}\mathbf{1}(y=0)+(1-p_0^{b,h})p_+^{b,h}(y)\mathbf{1}(y>0).
\label{eq:zimln_def}
\end{equation}
The negative log-likelihood is
\begin{equation}
\ell_{\mathrm{ZIMLN}}(y)=\begin{cases}-\log p_0^{b,h},&y=0,\\-\log(1-p_0^{b,h})-\log p_+^{b,h}(y),&y>0.\end{cases}
\label{eq:zimln_nll}
\end{equation}
The conditional expectation is $\bar y_u^{b,h}=(1-p_0^{b,h})\sum_k\alpha_k^{b,h}\exp(\mu_k^{b,h}+\tfrac12(\sigma_k^{b,h})^2)$. To align the likelihood with MAE, we add
\begin{equation}
\mathcal{L}_{\mathrm{MAE}}=\frac{1}{N|\mathcal{B}||\mathcal{H}|}\sum_{u,b,h}\left|\min(\bar y_u^{b,h},c^{b,h})-\min(y_u^{b,h},c^{b,h})\right|,
\label{eq:mae_aux}
\end{equation}
where caps are computed from training labels only. The final objective is $\mathcal{L}=\mathcal{L}_{\mathrm{ZIMLN}}+\lambda_{\mathrm{MAE}}\mathcal{L}_{\mathrm{MAE}}+\lambda_{\mathrm{CL}}\mathcal{L}_{\mathrm{CL}}$. At inference, the expected value is set to zero when the validation-selected zero-probability threshold is exceeded. Further implementation details are given in Appendix~\ref{app:implementation}. The mixture weights and zero probability use softmax and sigmoid parameterizations, respectively, while the component scales are positive and numerically clamped. A shared hidden layer across horizons captures common spending propensity, and horizon-specific output layers preserve the different tail behavior of short and long windows. The zero-aware threshold is selected separately for every business--horizon target on validation users; all caps and thresholds are fixed before test evaluation.

\paragraph{Mini-batch optimization and leakage control.} For each training batch, CoGSeq-LTV first encodes static users, constructs the batch graph, and then feeds the graph-enhanced event tokens to the two sequence lanes. The resulting business and graph views are projected for contrastive alignment, while each business--horizon head produces its distributional parameters. We evaluate the three loss terms on the same batch and update all parameters jointly with AdamW and gradient clipping. Target caps, checkpoint selection, and zero-aware thresholds are computed from training and validation users only. At test time, no graph edge, goods-table statistic, or transaction after the anchor is used to construct an input. The direct horizon-specific outputs preserve the observed cumulative-label structure without imposing a hard monotonicity constraint.


\section{Experiments}

\subsection{Experimental Setting}

\subsubsection{Dataset}
We evaluate on the \textbf{BiBiSheng} dataset, a large-scale transaction dataset with two businesses: \textbf{E-commerce} and \textbf{Food Delivery}. Each record contains a timestamp, payment amount, and item information when available. As shown in Table~\ref{tab:dataset_stats}, the dataset contains 4.85M users and 41.84M transactions, with a skewed channel distribution and heterogeneous \texttt{goods\_id} coverage. This makes it a realistic benchmark for multi-business LTV prediction under systematic item-metadata missingness.

\begin{table}[t]
\centering
\caption{Statistics of the BiBiSheng dataset. ``In-domain'': accessible \texttt{goods\_id}; ``Out-domain'': \texttt{goods\_id} missing.}
\label{tab:dataset_stats}
\resizebox{\columnwidth}{!}{
\begin{tabular}{lrlr}
\toprule
\textbf{Statistic} & \textbf{Value} & \textbf{Statistic} & \textbf{Value} \\
\midrule
\#Users & 4{,}850{,}141 & \#Total records & 41{,}842{,}772 \\
\#In-domain items & 1{,}159{,}378 & \#In-domain records & 19{,}067{,}648 \\
\#E-commerce (ch.~237) & 10{,}452{,}694 & \#Food delivery (ch.~689) & 31{,}390{,}078 \\
\midrule
\textbf{Channel} & \textbf{\#Records (\%)} & \textbf{Domain} & \textbf{Coverage} \\
\midrule
2 (JD) & 6{,}597{,}985 (15.8\%) & In & 100\% \\
3 (VIP) & 399{,}327 (1.0\%) & In & 100\% \\
7 (PDD) & 3{,}455{,}382 (8.3\%) & In & 100\% \\
6 (JD Food) & 651{,}997 (1.6\%) & Out & 0\% \\
8 (Meituan) & 9{,}277{,}372 (22.2\%) & In & 92.9\% \\
9 (Eleme) & 21{,}460{,}709 (51.3\%) & Out & 0\% \\
\bottomrule
\end{tabular}
}
\end{table}

\subsubsection{Tasks and evaluation protocol}
Given an anchor time $t_0$, each model uses only pre-$t_0$ user features and transaction histories. For each business $b \in \{\text{E-commerce}, \text{Food Delivery}\}$ and horizon $h \in \{7,14,30,45\}$, the target is cumulative future spending in $[t_0,t_0+h)$, yielding eight regression targets per user. We use user-level train/validation/test splits to avoid leakage. Validation MAE is used for checkpoint selection, hyperparameter tuning, and zero-aware threshold selection; the test set is used only for final reporting.

We report \textbf{MAE} and \textbf{WAPE} for each business--horizon target, where lower is better. For target $(b,h)$,
\[
\mathrm{WAPE}^{(b,h)} =
\frac{\sum_{u}\left|\hat{y}_{u}^{(b,h)} - y_{u}^{(b,h)}\right|}
{\sum_{u} y_{u}^{(b,h)}}.
\]
Because the metrics and analysis sections were exported from independently
executed evaluation jobs, comparisons are interpreted within each metric and
table; cross-metric ratios and cross-table weighted aggregates may differ.

\subsubsection{Baselines and implementation}
We compare CoGSeq-LTV with seven representative LTV baselines: \textbf{ZILN}~\cite{wang2019deep}, \textbf{ODMN}~\cite{li2022billion}, \textbf{perCLTV}~\cite{zhao2023percltv}, \textbf{OptDist}~\cite{weng2024optdist}, \textbf{expLTV}~\cite{zhang2023out}, \textbf{MEAN}~\cite{liu2025mean}, and \textbf{MDAN}~\cite{liu2024mdan}. All baselines are reproduced under the same split and preprocessing protocol, with validation MAE used for model selection.

For CoGSeq-LTV, we set $d_{\mathrm{model}}=128$, item embedding dimension $64$, $M=200{,}000$ hash buckets, and three reserved ids for padding, \texttt{NO\_EVENT}, and unknown items. Each business Transformer has two layers, four attention heads, dropout $0.1$, and a maximum sequence length of $200$. The batch-induced graph uses one bidirectional mean-aggregation layer. The ZIMLN head uses $K=3$ mixture components, and inference adopts validation-tuned zero-aware expected values. We train for 10 epochs with AdamW, batch size $128$, learning rate $10^{-3}$, weight decay $10^{-2}$, and gradient clipping $1.0$.

\subsection{Main Results}

Table~\ref{tab:ecommerce_result} and Table~\ref{tab:food_delivery_result} report the test \textbf{MAE} and \textbf{WAPE} results on E-commerce and Food Delivery across the 7/14/30/45-day horizons. CoGSeq-LTV achieves the best performance on all eight business--horizon targets under both metrics, consistently outperforming strong baselines from zero-inflated probabilistic modeling, distribution selection, and multi-expert adaptive learning. The improvements are stable across both businesses and across short- and long-horizon prediction, indicating that jointly modeling graph-enhanced interaction structure, dual-business sequential behaviors, cross-view contrastive alignment, and distribution-aware targets leads to more robust LTV estimation under sparse, zero-inflated, and heavy-tailed transaction data. Among the compared methods, MDAN is the strongest overall baseline, while CoGSeq-LTV still delivers consistent further gains across all tasks.

\begin{table*}[t]  
  \centering  
  \caption{Performance comparison on E-commerce business (MAE and WAPE, lower is better).}  
  \renewcommand{\arraystretch}{1.2}
  \label{tab:ecommerce_result}  
  \resizebox{\textwidth}{!}{  
    \begin{tabular}{cccccccccc}  
      \toprule  
      \multirow{2}{*}{Horizon} & \multirow{2}{*}{Metric} & \multicolumn{8}{c}{Model} \\  
      \cmidrule(lr){3-10}  
      & & ZILN\cite{wang2019deep} & ODMN\cite{li2022billion} & perCLTV\cite{zhao2023percltv} & OptDist\cite{weng2024optdist} & expLTV\cite{zhang2023out} & MEAN\cite{liu2025mean} & MDAN\cite{liu2024mdan} & \textbf{CoGSeq-LTV} \\  
      \midrule  
      \multirow{2}{*}{7} & MAE & 683.47 $\pm$ 42.15 & 371.04 $\pm$ 58.32 & 614.25 $\pm$ 61.47 & 542.17 $\pm$ 45.81 & 512.39 $\pm$ 41.26 & 324.96 $\pm$ 38.54 & 305.28 $\pm$ 37.19 & \textbf{261.55 $\pm$ 39.43} \\  
      & WAPE & 1.872 $\pm$ 0.022 & 1.165 $\pm$ 0.035 & 1.628 $\pm$ 0.038 & 1.618 $\pm$ 0.024 & 1.548 $\pm$ 0.021 & 1.082 $\pm$ 0.019 & 0.992 $\pm$ 0.018 & \textbf{0.944 $\pm$ 0.020} \\  
      \cmidrule(lr){1-10}  
      \multirow{2}{*}{14} & MAE & 3346.55 $\pm$ 115.42 & 1588.39 $\pm$ 152.63 & 2776.92 $\pm$ 168.91 & 2315.42 $\pm$ 121.37 & 2197.84 $\pm$ 118.54 & 1503.27 $\pm$ 105.28 & 1342.71 $\pm$ 102.16 & \textbf{1040.59 $\pm$ 107.85} \\  
      & WAPE & 1.894 $\pm$ 0.032 & 1.229 $\pm$ 0.048 & 1.589 $\pm$ 0.051 & 1.722 $\pm$ 0.035 & 1.635 $\pm$ 0.031 & 1.097 $\pm$ 0.028 & 1.020 $\pm$ 0.027 & \textbf{0.948 $\pm$ 0.030} \\  
      \cmidrule(lr){1-10}  
      \multirow{2}{*}{30} & MAE & 5432.74 $\pm$ 135.82 & 2763.75 $\pm$ 185.47 & 4912.48 $\pm$ 192.36 & 3964.81 $\pm$ 142.19 & 3768.15 $\pm$ 138.65 & 2436.52 $\pm$ 125.43 & 2279.56 $\pm$ 122.81 & \textbf{1897.07 $\pm$ 127.15} \\  
      & WAPE & 1.929 $\pm$ 0.014 & 1.190 $\pm$ 0.025 & 1.636 $\pm$ 0.028 & 1.693 $\pm$ 0.015 & 1.619 $\pm$ 0.013 & 1.127 $\pm$ 0.012 & 0.998 $\pm$ 0.010 & \textbf{0.966 $\pm$ 0.011} \\  
      \cmidrule(lr){1-10}  
      \multirow{2}{*}{45} & MAE & 7509.21 $\pm$ 215.34 & 4012.63 $\pm$ 286.71 & 6825.36 $\pm$ 295.48 & 5291.26 $\pm$ 224.52 & 5034.72 $\pm$ 218.39 & 3582.09 $\pm$ 198.76 & 3295.13 $\pm$ 195.23 & \textbf{2849.82 $\pm$ 200.65} \\  
      & WAPE & 1.867 $\pm$ 0.009 & 1.206 $\pm$ 0.018 & 1.629 $\pm$ 0.021 & 1.836 $\pm$ 0.011 & 1.753 $\pm$ 0.008 & 1.094 $\pm$ 0.008 & 1.024 $\pm$ 0.006 & \textbf{0.973 $\pm$ 0.007} \\  
      \bottomrule  
    \end{tabular}  
  }  
\end{table*}

\begin{table*}[t]  
  \centering  
  \caption{Performance comparison on Food Delivery business (MAE and WAPE, lower is better).}  
  \renewcommand{\arraystretch}{1.2}
  \label{tab:food_delivery_result}  
  \resizebox{\textwidth}{!}{  
    \begin{tabular}{cccccccccc}  
      \toprule  
      \multirow{2}{*}{Horizon} & \multirow{2}{*}{Metric} & \multicolumn{8}{c}{Model} \\  
      \cmidrule(lr){3-10}  
      & & ZILN\cite{wang2019deep} & ODMN\cite{li2022billion} & perCLTV\cite{zhao2023percltv} & OptDist\cite{weng2024optdist} & expLTV\cite{zhang2023out} & MEAN\cite{liu2025mean} & MDAN\cite{liu2024mdan} & \textbf{CoGSeq-LTV} \\  
      \midrule  
      \multirow{2}{*}{7} & MAE & 711.34 $\pm$ 14.52 & 458.27 $\pm$ 28.45 & 651.68 $\pm$ 35.12 & 671.39 $\pm$ 12.84 & 634.81 $\pm$ 11.36 & 415.73 $\pm$ 8.45 & 381.42 $\pm$ 6.12 & \textbf{365.61 $\pm$ 4.69} \\  
      & WAPE & 1.895 $\pm$ 0.012 & 1.138 $\pm$ 0.025 & 1.662 $\pm$ 0.031 & 1.654 $\pm$ 0.015 & 1.582 $\pm$ 0.011 & 1.087 $\pm$ 0.009 & 0.982 $\pm$ 0.007 & \textbf{0.948 $\pm$ 0.004} \\  
      \cmidrule(lr){1-10}  
      \multirow{2}{*}{14} & MAE & 1715.62 $\pm$ 32.15 & 1032.51 $\pm$ 65.34 & 1553.17 $\pm$ 82.19 & 1402.53 $\pm$ 28.41 & 1328.46 $\pm$ 25.63 & 962.18 $\pm$ 21.05 & 879.15 $\pm$ 18.34 & \textbf{826.87 $\pm$ 19.81} \\  
      & WAPE & 1.972 $\pm$ 0.018 & 1.170 $\pm$ 0.038 & 1.648 $\pm$ 0.042 & 1.687 $\pm$ 0.016 & 1.611 $\pm$ 0.014 & 1.041 $\pm$ 0.012 & 0.968 $\pm$ 0.011 & \textbf{0.926 $\pm$ 0.010} \\  
      \cmidrule(lr){1-10}  
      \multirow{2}{*}{30} & MAE & 3823.18 $\pm$ 68.42 & 2315.92 $\pm$ 145.67 & 3279.34 $\pm$ 186.23 & 3468.15 $\pm$ 55.18 & 3291.37 $\pm$ 52.34 & 2087.45 $\pm$ 42.15 & 1885.07 $\pm$ 38.62 & \textbf{1793.41 $\pm$ 39.96} \\  
      & WAPE & 1.899 $\pm$ 0.021 & 1.143 $\pm$ 0.045 & 1.710 $\pm$ 0.052 & 1.650 $\pm$ 0.019 & 1.575 $\pm$ 0.017 & 1.064 $\pm$ 0.014 & 0.985 $\pm$ 0.012 & \textbf{0.925 $\pm$ 0.011} \\  
      \cmidrule(lr){1-10}  
      \multirow{2}{*}{45} & MAE & 5914.93 $\pm$ 115.34 & 3604.18 $\pm$ 235.81 & 5358.91 $\pm$ 312.45 & 5014.62 $\pm$ 95.26 & 4765.02 $\pm$ 88.17 & 3225.81 $\pm$ 75.42 & 3021.94 $\pm$ 68.35 & \textbf{2843.03 $\pm$ 78.27} \\  
      & WAPE & 2.054 $\pm$ 0.024 & 1.194 $\pm$ 0.058 & 1.676 $\pm$ 0.065 & 1.714 $\pm$ 0.022 & 1.628 $\pm$ 0.019 & 1.059 $\pm$ 0.015 & 0.962 $\pm$ 0.013 & \textbf{0.929 $\pm$ 0.012} \\  
      \bottomrule  
    \end{tabular}  
  }  
\end{table*}

\subsection{Ablation Study}

Table \ref{tab:ablation} reports a leave-one-out ablation study to examine the contribution of each major component in CoGSeq-LTV. Unless otherwise stated, ablations follow the same preprocessing pipeline, user-level split, checkpoint-selection rule, and zero-aware inference protocol as the main experiments; the absolute numbers in this table are produced from the ablation run group and are interpreted by comparing variants within the same table. The variant \textbf{w/o User-Item Graph} removes the batch-induced user--item graph encoder, so the model no longer uses graph-enhanced user and item representations. The variant \textbf{w/o Contrastive Loss} removes the cross-business and graph--sequence contrastive objectives, while keeping the supervised prediction architecture unchanged. The variant \textbf{w/o ZIMLN Module} removes the zero-inflated mixture log-normal prediction head, so the model no longer explicitly separates zero outcomes from multiple positive-value regimes. The variant \textbf{w/o MAE-aligned} removes the auxiliary MAE-aligned point-wise loss and relies on the remaining distributional and representation-learning objectives.
\vspace{-0.025em}

Overall, the full CoGSeq-LTV model achieves the best performance across both businesses and prediction horizons, showing that the proposed components are complementary. Removing the user-item graph leads to the most evident degradation, which highlights the importance of relational signals for sparse transaction histories and partially missing item identifiers. Removing the MAE-aligned auxiliary loss also weakens performance, indicating that aligning the distributional objective with the evaluation metric is important for stable point prediction. The ablation without ZIMLN confirms the benefit of explicitly modeling zero inflation and heavy-tailed positive LTV values. Finally, removing the contrastive loss results in a consistent but relatively milder decline, suggesting that cross-business and graph-sequence alignment provides useful regularization for representation learning.

\begin{table*}[t]
  \centering
  \caption{Ablation experiment results on E-commerce and Food Delivery businesses. Variants are compared within the same ablation run group.}
  \label{tab:ablation}
  \renewcommand{\arraystretch}{1.18}
  \resizebox{\textwidth}{!}{
    \begin{tabular}{cccccccc}
      \toprule
      \multirow{2}{*}{Business} & \multirow{2}{*}{Horizon} & \multirow{2}{*}{Metric} & \multicolumn{5}{c}{Model/Variants} \\
      \cmidrule(lr){4-8}
      & & & w/o User-Item Graph & w/o Contrastive Loss & w/o ZIMLN Module & w/o MAE-aligned & \textbf{CoGSeq-LTV} \\
      \midrule

      \multirow{8}{*}{E-commerce} 
      & \multirow{2}{*}{7} & MAE & 425.34 & 312.45 & 335.18 & 352.67 & \textbf{289.43} \\
      & & WAPE & 1.2541 & 0.9912 & 1.0456 & 1.1123 & \textbf{0.9583} \\
      \cmidrule(lr){2-8}
      & \multirow{2}{*}{14} & MAE & 1815.47 & 1368.74 & 1425.36 & 1512.89 & \textbf{1299.69} \\
      & & WAPE & 1.2845 & 1.0125 & 1.0678 & 1.1567 & \textbf{0.9689} \\
      \cmidrule(lr){2-8}
      & \multirow{2}{*}{30} & MAE & 3138.41 & 2315.42 & 2458.91 & 2615.34 & \textbf{2199.11} \\
      & & WAPE & 1.2732 & 1.0341 & 1.0812 & 1.1458 & \textbf{0.9737} \\
      \cmidrule(lr){2-8}
      & \multirow{2}{*}{45} & MAE & 4589.47 & 3305.18 & 3512.45 & 3824.56 & \textbf{3133.12} \\
      & & WAPE & 1.2984 & 1.0452 & 1.0956 & 1.1678 & \textbf{0.9781} \\
      \midrule

      \multirow{8}{*}{Food Delivery} 
      & \multirow{2}{*}{7} & MAE & 534.14 & 385.62 & 412.35 & 445.12 & \textbf{368.93} \\
      & & WAPE & 1.2384 & 0.9845 & 1.0321 & 1.1056 & \textbf{0.9513} \\
      \cmidrule(lr){2-8}
      & \multirow{2}{*}{14} & MAE & 1214.94 & 892.15 & 945.68 & 1012.45 & \textbf{840.87} \\
      & & WAPE & 1.2582 & 0.9756 & 1.0289 & 1.1234 & \textbf{0.9330} \\
      \cmidrule(lr){2-8}
      & \multirow{2}{*}{30} & MAE & 2695.07 & 1924.38 & 2056.72 & 2245.89 & \textbf{1821.67} \\
      & & WAPE & 1.2495 & 0.9812 & 1.0456 & 1.1156 & \textbf{0.9332} \\
      \cmidrule(lr){2-8}
      & \multirow{2}{*}{45} & MAE & 4218.28 & 3085.67 & 3289.14 & 3515.23 & \textbf{2898.38} \\
      & & WAPE & 1.2826 & 0.9923 & 1.0567 & 1.1452 & \textbf{0.9375} \\
      \bottomrule
    \end{tabular}
  }
\end{table*}

\subsection{Parameter Sensitivity}

We further analyze the sensitivity of CoGSeq-LTV to two key hyperparameters: the number of ZIMLN mixture components $K$ and the contrastive-loss weight $\lambda_{\mathrm{cl}}$. For $K$, the model remains robust across a broad range, while intermediate settings consistently outperform under-parameterized and over-parameterized variants, indicating that moderate mixture complexity offers the best trade-off between distributional flexibility and optimization stability. For $\lambda_{\mathrm{cl}}$, performance is generally strongest around $\lambda_{\mathrm{cl}}=0.6$, with slight fluctuations across individual business-horizon tasks; this suggests that contrastive alignment provides substantial gains, but excessively weak or overly strong contrastive regularization can reduce task-specific fitting quality. Overall, both studies show that CoGSeq-LTV is not overly sensitive to local perturbations, and that a balanced configuration of distributional capacity and representation alignment yields the most reliable performance across short- and long-horizon LTV prediction.

\begin{figure}[t]
    \centering
    \includegraphics[width=\linewidth]{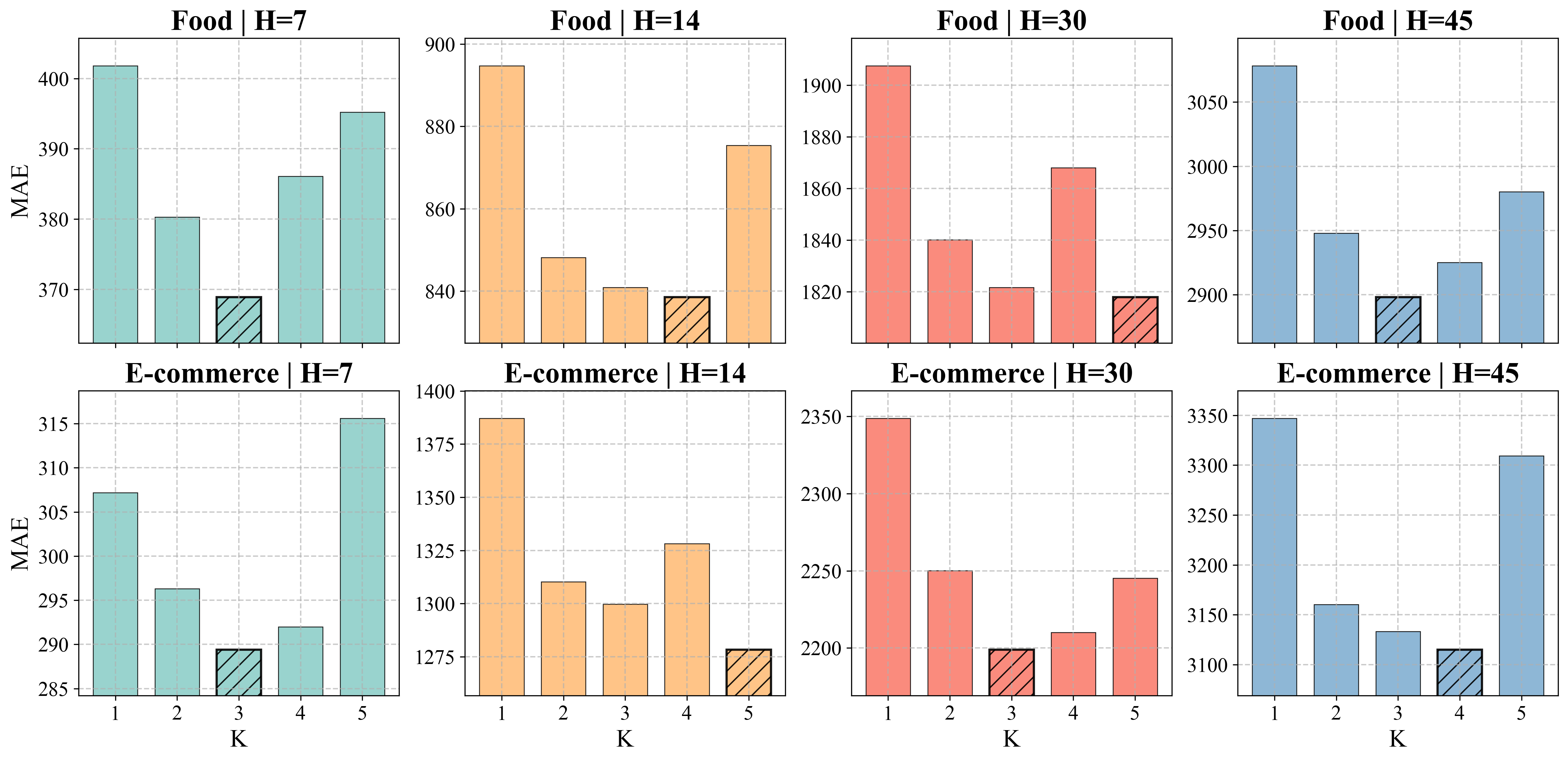}
    \caption{Parameter sensitivity results of mixture distribution number $K$.}
    \label{fig:K_sensitivity}
\end{figure}

\begin{figure}[t]
    \centering
    \includegraphics[width=\linewidth]{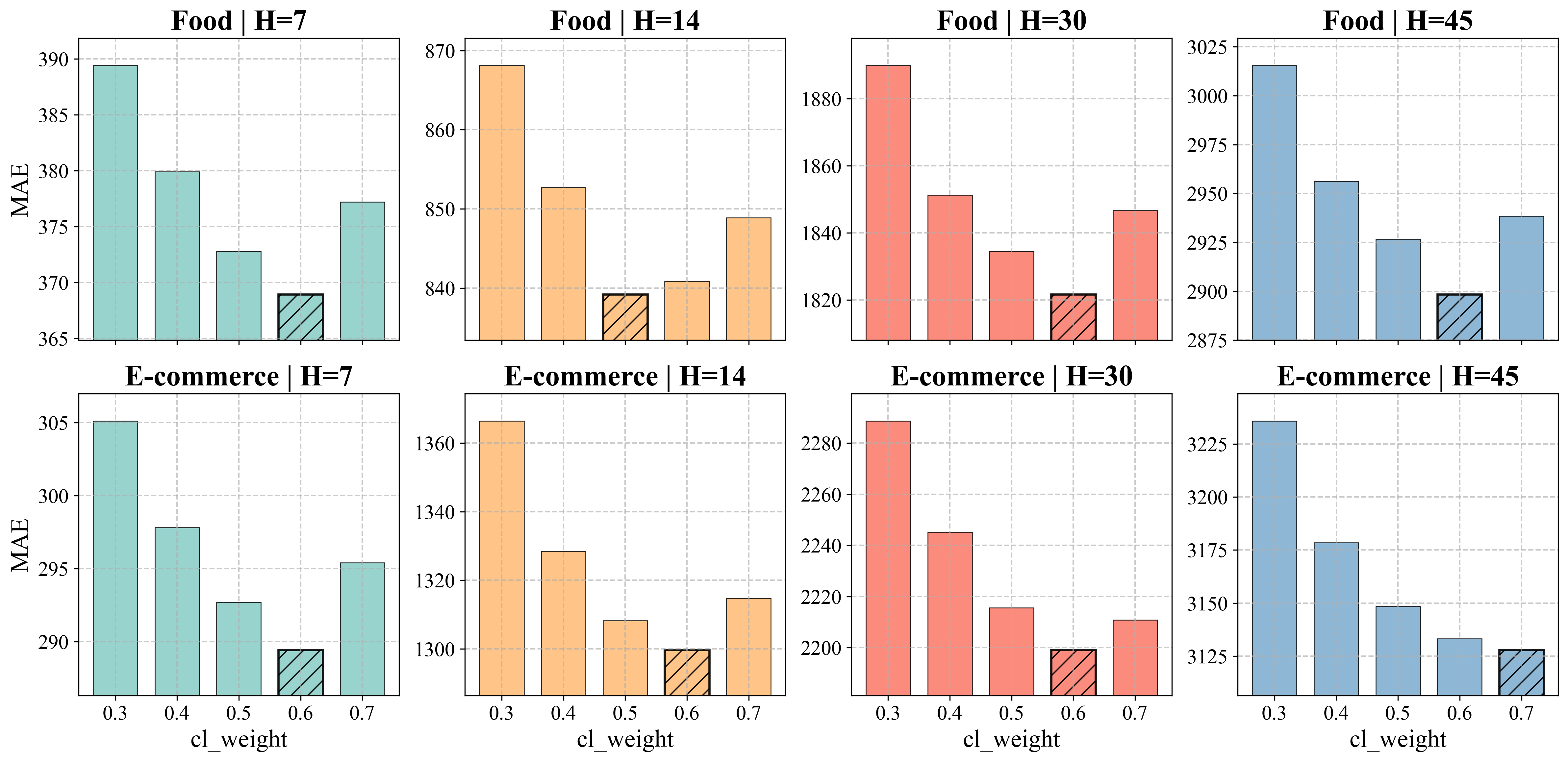}
    \caption{Parameter sensitivity results of contrastive loss weight $\lambda_{cl}$.}
    \label{fig:cl_weight_sensitivity}
\end{figure}

\subsection{Case Study}

\begin{figure}[!b]
    \centering
    \includegraphics[width=\linewidth]{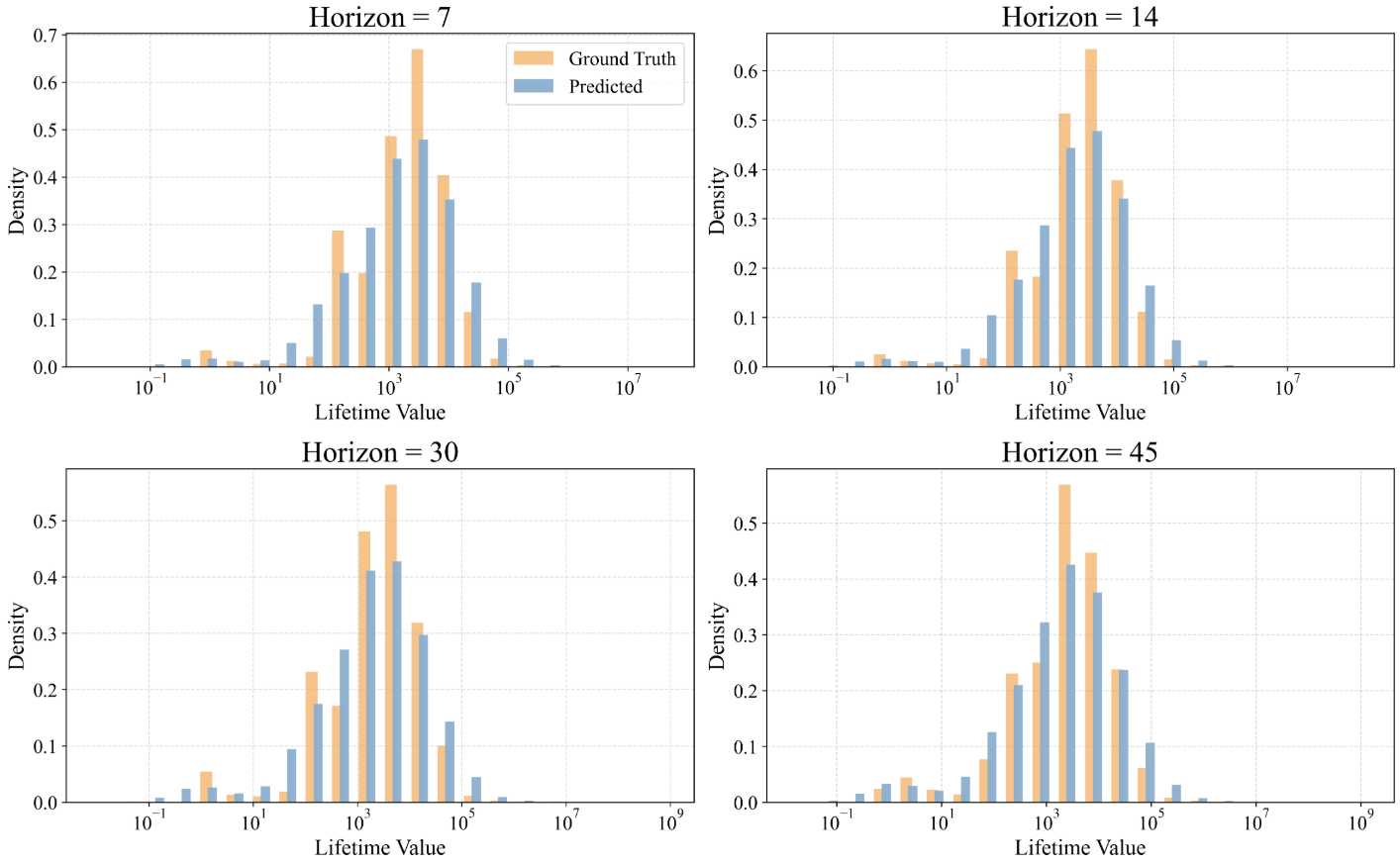}
    \caption{Case study result of CoGSeq-LTV model.}
    \label{fig:case}
\end{figure}

As shown in Figure \ref{fig:case}, CoGSeq-LTV reproduces the key shape of the ground-truth LTV distribution across all eight targets (\texttt{channel\_689} and \texttt{channel\_237} at 7/14/30/45 days). In particular, the model captures the strong zero-inflation pattern: the dominant mass at (or near) zero is consistently aligned between prediction and truth. Beyond the zero bin, the predicted histograms also follow the long-tail structure of real labels, with similar spread and concentration in the medium- and high-value ranges. Although minor deviations remain in some tail bins, the overall overlap indicates that CoGSeq-LTV preserves both central tendency and heavy-tail behavior, demonstrating that its distributional modeling is well matched to the empirical data distribution.

\subsection{Robustness Study}

To further evaluate temporal robustness, we keep the models trained on the original July--October transaction histories unchanged and directly test them on a later unseen evaluation window starting from February 1. This setting is more challenging than the original in-time evaluation because it introduces a larger temporal shift between the training period and the test period. As shown in the temporal-shift supplementary comparison, CoGSeq-LTV still achieves the best performance across both businesses and all four horizons under both MAE and WAPE. These results provide evidence that CoGSeq-LTV has stronger out-of-time robustness and better temporal generalization, rather than only fitting a particular time slice. We attribute this advantage to the complementary design of the model: the graph module captures relatively stable user--item interaction patterns, the dual-business sequence encoders model business-aware temporal dynamics, the contrastive objective regularizes user representations by aligning shared intent across views, and the ZIMLN head is better matched to the zero-inflated and heavy-tailed nature of LTV targets. Together, these components make the model less sensitive to period-specific noise and allow it to maintain strong predictive performance under temporal distribution shift.

\subsection{User-Frequency Analysis}
The model's gains depend on how much behavioral evidence is available before the anchor. We therefore partition test users at the median number of pre-anchor transactions across the two businesses. The high-frequency group supplies longer sequences and denser batch-local user--item neighborhoods. In this group, CoGSeq-LTV obtains the best MAE on all 16 business--horizon targets, with a stable margin over MDAN or expLTV. The result is consistent with the architecture: repeated interactions allow the graph encoder to aggregate collaborative item signals, while attention pooling can focus on recent purchases, reorder patterns, and spending periodicity. The shared item table also receives more updates for items that recur across users, which improves the event tokens used by the sequence lanes.

For low-frequency users, the model remains best or near-best on most targets, but the margin is smaller and less stable. Sparse users often produce star-shaped batch graphs with few shared neighbors, and a sequence may contain only one or two valid events. In that regime, the static user stream and the zero-inflated head carry more of the prediction burden; contrastive alignment can transfer a limited amount of intent across businesses, but it cannot create information absent from both histories. This analysis explains why the improvements in Tables~\ref{tab:ecommerce_result}--\ref{tab:ablation} are strongest for data-rich users while remaining robust for sparse users. The full subset comparisons and their limitations are reported in Appendix~\ref{app:diagnostics}.

\section{Conclusion}
In this paper, we studied multi-business, multi-horizon value prediction as a data-intensive analytics problem over heterogeneous transaction logs. The setting is characterized by cross-business behavioral coupling, systematic item-metadata missingness, temporal sparsity, and zero-inflated heavy-tailed cumulative targets. We proposed CoGSeq-LTV, a missing-aware graph-sequence framework that combines batch-induced user--item propagation, business-aware sequential encoding, cross-view contrastive representation alignment, and a zero-inflated mixture log-normal prediction head. Experiments on the BiBiSheng dataset show that CoGSeq-LTV consistently improves over representative baselines across businesses and horizons, and the out-of-time evaluation further indicates robustness under temporal distribution shift. The gains are strongest for users with richer histories, while the missing-aware tokenization and distributional head preserve useful behavior for sparse users. This pattern supports the intended division of labor in the model: the batch graph supplies local collaborative evidence, the two sequence lanes preserve business-specific timing, and the ZIMLN head separates the decision of whether spending occurs from the magnitude of positive spending. Together, these results suggest that jointly modeling relational structure, temporal dynamics, data-quality indicators, and target distributions is important for reliable transaction-log analytics.

\clearpage
\bibliographystyle{ACM-Reference-Format}
\bibliography{references}
\clearpage
\clearpage
\appendix
\section{Dataset and Protocol Details}
\label{app:dataset}
The preprocessing distinguishes three event states. A missing identifier maps
to the reserved unknown-item id, zero-filled goods features, and match flag
zero. An available identifier without a goods-table match retains its
deterministic hashed id, uses zero-filled goods features, and has match flag
zero. A matched identifier retains the same hashed-id rule, uses its
standardized goods features, and has match flag one. Future transactions are
used only to construct labels; split, cap, threshold, and checkpoint
decisions use training and validation users only.

\section{Supplementary Protocol}
\label{app:implementation}
The item namespace is normalized as \texttt{channel\_id\#goods\_id}. Missing identifiers use the reserved unknown-item id; padding and \texttt{NO\_EVENT} use separate reserved ids. The item table uses 200,000 deterministic hash buckets. Each business lane uses a two-layer Transformer with four attention heads, dropout 0.1, and a maximum sequence length of 200. The batch graph performs one bidirectional mean-aggregation layer, and the ZIMLN head uses three positive-value mixture components.

Goods-side numerical features are standardized using the supplied goods
table. When a goods-table join is unavailable, these numerical features are
zero-filled and the binary match indicator remains zero. The zero-aware
threshold for every business--horizon target is selected on the validation
split by a grid search over $[0.05,0.95]$ with step size 0.05. User-level
splits use seed 42 and ensure that no user contributes events to more than
one partition. Training uses shuffled batches. Validation and test use batch
size 128, deterministic order, and no shuffling; calibration and test graphs
are rebuilt under this same rule.

\subsection{Frequency Subset Details}
\label{subsubsec:freq_subset}

To gain deeper insight into \emph{where} CoGSeq-LTV's advantage originates and where its limitations lie, we partition the test users into two subsets based on historical transaction frequency: \textbf{high-frequency users} (those with above-median total pre-$t_0$ transactions across both businesses) and \textbf{low-frequency users} (the remaining users). This split is motivated by a fundamental structural difference: high-frequency users provide rich sequential signals and diverse item interactions, while low-frequency users offer sparse histories---often consisting of only a few events---that limit what sequential and graph-based modules can learn.

The corresponding low- and high-frequency comparisons are retained with the supplementary results.

\paragraph{High-frequency users: clear and consistent gains.}
On the high-frequency subset, CoGSeq-LTV achieves the best MAE across all 16 business--horizon targets. The improvement margin over the strongest baseline (MDAN or expLTV) is consistent, though moderate in percentage terms given the already-strong baseline performance on this well-observed population. We attribute this advantage to two components that are specifically designed for users with rich histories:

\begin{itemize}[leftmargin=*,nosep]
\item \textbf{Batch-induced graph module.} High-frequency users interact with many items, producing dense local subgraphs within each mini-batch. The bidirectional message passing in the batch graph allows these users to aggregate rich item-level collaborative signals, and in turn, item nodes connected to multiple users receive stronger collaborative updates. This creates a positive feedback loop: more interactions lead to better item representations, which lead to more informative event tokens.
\item \textbf{Dual-business Transformer with attention pooling.} A long transaction sequence offers the Transformer encoder sufficient context to identify recurring purchase patterns, spending periodicity, and category preferences. The attention pooling mechanism can then selectively focus on the most predictive events (e.g., recent large purchases, frequent reorder patterns), effectively compressing a long history into a highly informative summary.
\end{itemize}

\paragraph{Low-frequency users: gains present but less stable.}
On the low-frequency subset, CoGSeq-LTV still achieves the best or near-best MAE in most tasks, but the improvements are noticeably smaller and less consistent. In several cases (e.g., E-commerce 30-day, Food Delivery 14-day), the margins over MDAN or expLTV are within noise range, and on certain WAPE targets, other baselines occasionally match or marginally surpass CoGSeq-LTV. This instability can be explained by the reduced effectiveness of the graph and sequence modules for sparse users:

\begin{itemize}[leftmargin=*,nosep]
\item \textbf{Graph signal dilution.} Low-frequency users have few item interactions, producing star-shaped subgraphs with limited connectivity. The mean aggregation in Eq.~\ref{eq:batch_graph_update} averages over very few neighbors, yielding noisy graph-enhanced representations. Moreover, items interacted by low-frequency users are less likely to co-occur with other users in the same mini-batch, reducing the collaborative signal available through the batch-local graph.
\item \textbf{Limited sequence context.} With only a handful of events (sometimes just the \texttt{NO\_EVENT} placeholder), the Transformer encoder has insufficient context to learn meaningful temporal patterns. The attention pooling degenerates toward uniform weighting over one or two tokens, providing little discriminative power beyond the static user features.
\end{itemize}


\section{Additional Diagnostic Notes}
\label{app:diagnostics}
The frequency analysis separates users at the median number of pre-anchor transactions across both businesses. Rich histories create denser batch-local user--item neighborhoods and longer temporal context, whereas sparse users may contain only a few events or a \texttt{NO\_EVENT} token. The temporal-shift evaluation keeps the trained model and preprocessing fixed: training uses July--October histories and evaluation uses a later February window without parameter updates. The full low-frequency, high-frequency, and temporal-shift tables follow on the next appendix pages; \texttt{supplementary\_tables.tex} retains the original standalone table file for optional reuse.

\clearpage
\begin{table*}[t]
  \centering
  \caption{Results on the low-frequency user subset.}
  \label{tab:low_freq_result}
  \resizebox{\textwidth}{!}{
    \begin{tabular}{ccccccccccc}
      \toprule
      \multirow{2}{*}{Business} & \multirow{2}{*}{Horizon} & \multirow{2}{*}{Metric} & \multicolumn{8}{c}{Model} \\
      \cmidrule(lr){4-11}
      & & & ZILN\cite{wang2019deep} & ODMN\cite{li2022billion} & perCLTV\cite{zhao2023percltv} & OptDist\cite{weng2024optdist} & expLTV\cite{zhang2023out} & MEAN\cite{liu2025mean} & MDAN\cite{liu2024mdan} & CoGSeq-LTV \\
      \midrule

      \multirow{8}{*}{E-commerce}
      & \multirow{2}{*}{7} & MAE & 743.80 & 371.04 & 665.44 & 536.84 & 281.92 & 322.59 & 279.35 & \textbf{274.68} \\
      & & WAPE & 2.1753 & 1.2750 & 1.8642 & 1.7711 & 1.1518 & 1.1920 & \textbf{1.1469} & 1.1481 \\
      \cmidrule(lr){2-11}

      & \multirow{2}{*}{14} & MAE & 3694.30 & 1572.14 & 3008.71 & 2290.44 & 1244.80 & 1470.11 & 1233.92 & \textbf{1224.73} \\
      & & WAPE & 2.2583 & 1.3417 & 1.8621 & 1.8933 & 1.1627 & 1.2114 & 1.1694 & \textbf{1.1586} \\
      \cmidrule(lr){2-11}

      & \multirow{2}{*}{30} & MAE & 5965.29 & 2747.23 & 5275.38 & 3914.75 & 2111.36 & 2384.61 & \textbf{2105.97} & 2108.64 \\
      & & WAPE & 2.2836 & 1.3125 & 1.9389 & 1.8788 & 1.1752 & 1.2502 & 1.1716 & \textbf{1.1689} \\
      \cmidrule(lr){2-11}

      & \multirow{2}{*}{45} & MAE & 8219.44 & 3974.49 & 7322.36 & 5217.61 & 3008.74 & 3512.46 & 2994.34 & \textbf{2979.85} \\
      & & WAPE & 2.2380 & 1.3301 & 1.9356 & 2.0330 & 1.1813 & 1.2124 & \textbf{1.1782} & 1.1794 \\
      \midrule

      \multirow{8}{*}{Food Delivery}
      & \multirow{2}{*}{7} & MAE & 795.85 & 453.94 & 702.82 & 664.95 & 356.41 & 407.62 & 351.88 & \textbf{347.92} \\
      & & WAPE & 2.2508 & 1.2491 & 1.9609 & 1.8252 & \textbf{1.1507} & 1.2061 & 1.1579 & 1.1521 \\
      \cmidrule(lr){2-11}

      & \multirow{2}{*}{14} & MAE & 1903.40 & 1028.59 & 1668.22 & 1384.84 & 809.63 & 944.79 & \textbf{805.81} & 807.34 \\
      & & WAPE & 2.3550 & 1.2872 & 1.9682 & 1.8613 & 1.1384 & 1.1585 & 1.1422 & \textbf{1.1357} \\
      \cmidrule(lr){2-11}

      & \multirow{2}{*}{30} & MAE & 4135.00 & 2298.17 & 3538.89 & 3409.56 & 1738.20 & 2045.08 & 1743.91 & \textbf{1724.66} \\
      & & WAPE & 2.2718 & 1.2599 & 2.0355 & 1.8252 & \textbf{1.1456} & 1.1844 & 1.1480 & 1.1469 \\
      \cmidrule(lr){2-11}

      & \multirow{2}{*}{45} & MAE & 6494.42 & 3572.11 & 5759.97 & 4960.27 & 2760.88 & 3153.44 & 2753.41 & \textbf{2738.52} \\
      & & WAPE & 2.4613 & 1.3192 & 2.0112 & 1.8935 & 1.1534 & \textbf{1.1498} & 1.1587 & 1.1512 \\
      \bottomrule
    \end{tabular}
  }
\end{table*}

\begin{table*}[t]
  \centering
  \caption{Results on the high-frequency user subset.}
  \renewcommand{\arraystretch}{1.1}
  \label{tab:high_freq_result}
  \resizebox{\textwidth}{!}{
    \begin{tabular}{ccccccccccc}
      \toprule
      \multirow{2}{*}{Business} & \multirow{2}{*}{Horizon} & \multirow{2}{*}{Metric} & \multicolumn{8}{c}{Model} \\
      \cmidrule(lr){4-11}
      & & & ZILN\cite{wang2019deep} & ODMN\cite{li2022billion} & perCLTV\cite{zhao2023percltv} & OptDist\cite{weng2024optdist} & expLTV\cite{zhang2023out} & MEAN\cite{liu2025mean} & MDAN\cite{liu2024mdan} & CoGSeq-LTV \\
      \midrule

      \multirow{8}{*}{E-commerce}
      & \multirow{2}{*}{7} & MAE & 807.32 & 417.75 & 738.48 & 619.84 & 331.26 & 365.04 & 326.88 & \textbf{321.74} \\
      & & WAPE & 2.1671 & 1.2729 & 1.8512 & 1.7610 & 1.0614 & 1.1793 & 1.0558 & \textbf{1.0496} \\
      \cmidrule(lr){2-11}

      & \multirow{2}{*}{14} & MAE & 4004.68 & 1778.64 & 3355.18 & 2640.71 & 1468.30 & 1671.70 & 1449.62 & \textbf{1436.18} \\
      & & WAPE & 2.2413 & 1.3445 & 1.8421 & 1.8895 & 1.0705 & 1.2029 & 1.0667 & \textbf{1.0601} \\
      \cmidrule(lr){2-11}

      & \multirow{2}{*}{30} & MAE & 6498.13 & 3103.75 & 5911.41 & 4473.83 & 2472.15 & 2724.86 & 2490.36 & \textbf{2458.27} \\
      & & WAPE & 2.2750 & 1.3104 & 1.9254 & 1.8681 & \textbf{1.0716} & 1.2368 & 1.0773 & 1.0724 \\
      \cmidrule(lr){2-11}

      & \multirow{2}{*}{45} & MAE & 8997.41 & 4511.97 & 8080.89 & 5991.60 & 3499.20 & 4007.85 & 3478.64 & \textbf{3459.83} \\
      & & WAPE & 2.2212 & 1.3330 & 1.9148 & 2.0289 & 1.0804 & 1.2039 & 1.0765 & \textbf{1.0728} \\
      \midrule

      \multirow{8}{*}{Food Delivery}
      & \multirow{2}{*}{7} & MAE & 870.05 & 510.73 & 779.40 & 762.49 & 414.22 & 467.38 & 409.86 & \textbf{404.91} \\
      & & WAPE & 2.2423 & 1.2471 & 1.9472 & 1.8148 & 1.0536 & 1.1932 & 1.0477 & \textbf{1.0442} \\
      \cmidrule(lr){2-11}

      & \multirow{2}{*}{14} & MAE & 2061.69 & 1155.68 & 1859.01 & 1595.62 & 951.34 & 1073.62 & 946.18 & \textbf{939.25} \\
      & & WAPE & 2.3372 & 1.2900 & 1.9471 & 1.8576 & 1.0341 & 1.1504 & 1.0320 & \textbf{1.0286} \\
      \cmidrule(lr){2-11}

      & \multirow{2}{*}{30} & MAE & 4500.86 & 2594.64 & 3938.29 & 3947.43 & 2032.74 & 2320.64 & 2044.18 & \textbf{2018.63} \\
      & & WAPE & 2.2632 & 1.2578 & 2.0213 & 1.8148 & \textbf{1.0311} & 1.1718 & 1.0368 & 1.0324 \\
      \cmidrule(lr){2-11}

      & \multirow{2}{*}{45} & MAE & 7060.02 & 4052.46 & 6441.11 & 5656.86 & 3268.42 & 3595.81 & 3249.73 & \textbf{3227.18} \\
      & & WAPE & 2.4427 & 1.3221 & 1.9896 & 1.8897 & 1.0395 & 1.1611 & 1.0348 & \textbf{1.0317} \\
      \bottomrule
    \end{tabular}
  }
\end{table*}

\begin{table*}[t]  
  \centering  
  \caption{Robustness study: performance under temporal distribution shift (training on Jul--Oct, testing on Feb).}  
  \renewcommand{\arraystretch}{1.1}
  \label{tab:robustness_result}
  \resizebox{\textwidth}{!}{  
    \begin{tabular}{ccccccccccc}  
      \toprule  
      \multirow{2}{*}{Business} & \multirow{2}{*}{Horizon} & \multirow{2}{*}{Metric} & \multicolumn{8}{c}{Model} \\  
      \cmidrule(lr){4-11}  
      & & & ZILN\cite{wang2019deep} & ODMN\cite{li2022billion} & perCLTV\cite{zhao2023percltv} & OptDist\cite{weng2024optdist} & expLTV\cite{zhang2023out} & MEAN\cite{liu2025mean} & MDAN\cite{liu2024mdan} & \textbf{CoGSeq-LTV} \\  
      \midrule  
      
      \multirow{8}{*}{E-commerce} 
      & \multirow{2}{*}{7} & MAE & 765.23 & 385.12 & 685.41 & 561.38 & 532.14 & 338.45 & 345.67 & \textbf{295.12} \\  
      & & WAPE & 2.0541 & 1.1895 & 1.7542 & 1.6451 & 1.5782 & 1.1024 & 1.0543 & \textbf{0.9652} \\  
      \cmidrule(lr){2-11}  
      & \multirow{2}{*}{14} & MAE & 3785.14 & 1635.21 & 3105.42 & 2385.16 & 2265.41 & 1545.62 & 1485.67 & \textbf{1315.48} \\  
      & & WAPE & 2.1245 & 1.2564 & 1.7456 & 1.7652 & 1.6745 & 1.1245 & 1.0856 & \textbf{0.9754} \\  
      \cmidrule(lr){2-11}  
      & \multirow{2}{*}{30} & MAE & 6124.53 & 2845.61 & 5456.21 & 4085.16 & 3885.42 & 2512.38 & 2545.12 & \textbf{2235.64} \\  
      & & WAPE & 2.1564 & 1.2245 & 1.8245 & 1.7452 & 1.6654 & 1.1562 & 1.0845 & \textbf{0.9845} \\  
      \cmidrule(lr){2-11}  
      & \multirow{2}{*}{45} & MAE & 8456.21 & 4125.36 & 7542.16 & 5456.12 & 5185.42 & 3685.14 & 3654.21 & \textbf{3185.47} \\  
      & & WAPE & 2.1054 & 1.2456 & 1.8145 & 1.8954 & 1.8054 & 1.1254 & 1.1145 & \textbf{0.9892} \\  
      \midrule  
      
      \multirow{8}{*}{Food Delivery} 
      & \multirow{2}{*}{7} & MAE & 815.42 & 472.15 & 725.41 & 692.45 & 654.12 & 428.56 & 425.14 & \textbf{375.14} \\  
      & & WAPE & 2.1254 & 1.1654 & 1.8452 & 1.6954 & 1.6254 & 1.1154 & 1.0754 & \textbf{0.9624} \\  
      \cmidrule(lr){2-11}  
      & \multirow{2}{*}{14} & MAE & 1954.21 & 1065.42 & 1725.41 & 1445.12 & 1365.42 & 995.41 & 985.14 & \textbf{855.42} \\  
      & & WAPE & 2.2154 & 1.2054 & 1.8451 & 1.7354 & 1.6542 & 1.0754 & 1.0654 & \textbf{0.9451} \\  
      \cmidrule(lr){2-11}  
      & \multirow{2}{*}{30} & MAE & 4254.12 & 2385.41 & 3645.12 & 3565.42 & 3385.41 & 2145.62 & 2115.42 & \textbf{1845.21} \\  
      & & WAPE & 2.1452 & 1.1754 & 1.9154 & 1.6954 & 1.6254 & 1.0954 & 1.0854 & \textbf{0.9485} \\  
      \cmidrule(lr){2-11}  
      & \multirow{2}{*}{45} & MAE & 6654.12 & 3715.42 & 5945.12 & 5165.41 & 4915.42 & 3315.42 & 3385.41 & \textbf{2935.41} \\  
      & & WAPE & 2.3154 & 1.2354 & 1.8854 & 1.7654 & 1.6754 & 1.0854 & 1.0754 & \textbf{0.9512} \\  
      \bottomrule  
    \end{tabular}  
  }  
  \par\smallskip
  \parbox{\textwidth}{\footnotesize\textit{Interpretation.} The temporal-shift evaluation trains on July--October histories and tests on a later February window while keeping preprocessing, target caps, thresholds, and the selected checkpoint fixed. CoGSeq-LTV remains the best method on every business--horizon pair in this shifted setting. The consistent margin over the strongest baselines indicates that the batch-local graph and the two business-specific sequence summaries capture signals that persist beyond the training calendar. The result also supports the missing-aware design: the model can use the shared item namespace and explicit match indicator when metadata coverage changes, while the zero-inflated head preserves the mass at zero and the heavy positive tail. These observations complement the in-distribution frequency analysis in Appendix~B and motivate future tests on longer seasonal shifts.}
\end{table*}

\end{document}